\documentclass[review]{elsarticle}

\usepackage{amsmath}
\usepackage{algorithm}
\usepackage{algorithmic}
\usepackage{multirow}
\usepackage{booktabs}
\usepackage{graphicx}

\journal{Journal of \LaTeX\ Templates}

\begin{document}

\begin{frontmatter}

\title{Multi-fidelity surrogate modeling for temperature field prediction using deep convolution neural network}

\author[1]{Yunyang Zhang}
\author[1]{Zhiqiang Gong}
\author[1]{Weien Zhou}
\author[1]{Xiaoyu Zhao}
\author[2]{Xiaohu Zheng}
\author[1]{Wen Yao\corref{mycorrespondingauthor}}
\cortext[mycorrespondingauthor]{Corresponding author}
\address[1]{Defense Innovation Institute,
	Chinese Academy of Military Science, Beijing, China\\}
\address[2]{College of Aerospace Science and Engineering,
	National University of Defense Technology, Changsha, China\\}

\begin{abstract}
Temperature field prediction is of great importance in the thermal design of systems engineering, and building the surrogate model is an effective way for the task.
Generally, large amounts of labeled data are required to guarantee a good prediction performance of the surrogate model, especially the deep learning model, which have more parameters and better representational ability. 
However, labeled data, especially high-fidelity labeled data, are usually expensive to obtain and sometimes even impossible.
To solve this problem, this paper proposes a pithy deep multi-fidelity model (DMFM) for temperature field prediction, which takes advantage of low-fidelity data to boost the performance with less high-fidelity data.
First, a pre-train and fine-tune paradigm are developed in DMFM to train the low-fidelity and high-fidelity data, which significantly reduces the complexity of the deep surrogate model.
Then, a self-supervised learning method for training the physics-driven deep multi-fidelity model (PD-DMFM) is proposed,  which fully utilizes the physics characteristics of the engineering systems and reduces the dependence on large amounts of labeled low-fidelity data in the training process.
Two diverse temperature field prediction problems are constructed to validate the effectiveness of DMFM and PD-DMFM, and the result shows that the proposed method can greatly reduce the dependence of the model on high-fidelity data.
\end{abstract}


\end{frontmatter}


\section{Introduction}

With the development of electronic equipment towards miniaturization and high performance, heat dissipation has gradually become the main factor affecting product performance and quality \cite{Yao2011}. High-performance electronic devices with high power generate a lot of heat in the work. Meanwhile, miniaturization leads to limited heat dissipation, further bringing out heat accumulation of electronic equipment, harming the performance and the service life of equipment, and even causing serious accidents. It is necessary to control the temperature of electronic components in the product by thermal design so as to guarantee the safety and reliability of the devices \cite{Zheng2021,Zheng2019,Zheng2020,he2021thermal,ahmed2018optimization}.

Accurate temperature prediction of electronic equipment is the premise of thermal design. Building a surrogate model to predict the temperature fields of equipment is an effective method that can significantly save the cost of numerical calculation. Typical surrogate modeling methods include polynomial response surface (PRS) \cite{goel2009comparing}, Kriging \cite{clark2016engineering}, radial basis function (RBF) \cite{yao2011concurrent}, and support vector regression (SVR) \cite{clarke2005analysis}. 
Above surrogate modeling methods can learn the model adaptively and thus could capture the physical information and provide the temperature field information.
However, due to the limited representation ability of the model, especially in the high-dimensional prediction case such as temperature field prediction, these methods usually cannot meet the performance requirements in engineering systems.
In recent years, deep learning (DL), which has powerful potential in solving high-dimensional and strongly nonlinear problems, has provided an alternative way for surrogate modeling \cite{lecun2015deep,chen2021deep,zhao2021physics,chen2020heat,zhao2021surrogate}. 
The deep learning model represented by the convolutional neural network (CNN) \cite{gong2019cnn, gong2020statistical} can extract both the local and global physical information of the system and thus provide good performance on temperature field prediction. \textbf{This paper uses the deep convolutional neural network as the surrogate model for temperature field prediction.}

The success of deep learning depends on a large number of labeled data. 
The dataset used in computer vision, such as Imagenet, usually contains millions of labeled samples. 
For the temperature field prediction task, the labeled data is usually obtained through physical experiments or numerical simulation, requiring amount cost of labor and time. 
For example, the numerical thermal simulation of complex electronic devices usually takes several hours or even days. Producing thousands labeled data for deep surrogate model training often takes months, which heavily delays the thermal design cycle of electronic devices. 
Besides, it is hard to obtain the ground truth of the temperature field in some special cases, such as the on-orbit satellite. Therefore, to practical apply deep learning as the surrogate model of an engineering system, \textbf{it is necessary to solve the defect of the prohibitive cost of labeled data acquisition.} 

The multi-fidelity modeling method exploring abundant low-fidelity (LF) data and limited high-fidelity (HF) data is considered the potential method to effectively reduce the acquisition cost of labeled data \cite{peherstorfer2018survey,fernandez2016review}. The high-fidelity data are often obtained through physical experiments or high-precision simulations, which are scarce and expensive but accurate. While, the low-fidelity data are usually acquired by low-precision simulations, which are cheap and easy to obtain but inaccurate. Although inaccurate, the low-fidelity data contain useful information, roughly reflecting the trend of high-fidelity data. 
The multi-fidelity model trained through low-fidelity and high-fidelity data could take advantage of the merits of two classes of data and provide efficient and accurate performance on temperature field prediction. 
Traditional multi-fidelity modeling methods, including Co-kriging \cite{le2014recursive,perdikaris2015multi} and multi-level Monte Carle (MLMO) \cite{bierig2016approximation, giles2008multilevel}, have been proven to be successful and effective in reducing the amount of required high-fidelity data while maintaining the high precision of the model. 
In recent years, some works have attempted to utilize multi-fidelity modeling on deep neural networks \cite{liu2019multi,zhang2021multi,chen2022multi,song2021transfer,meng2020composite}. 
Liu et al. \cite{liu2019multi} first build a base model employing the low fidelity data and then train another neural network with limited high-fidelity data to predict differences between low-fidelity and high-fidelity data. During the inference, the two models need to be jointly employed.
Meng et al. \cite{meng2020composite} considers the complex mapping between low-fidelity and high-fidelity data and use two other networks to fit the linear and nonlinear relationship between low-fidelity and high-fidelity data based on the low-fidelity model. 
Chen et al. \cite{chen2022multi} explore the relationship between high-fidelity data and every low-fidelity datum, make full use of a large number of low-fidelity data.
The above methods utilize multiple neural networks to achieve multi-fidelity modeling and need to combine numerous networks for inferencing in practical use. On the one hand, the structure of the serial multi-fidelity model is redundant; on the other hand, the series of multiple models may cause error accumulation and propagation.
In addition, the existing deep multi-fidelity surrogate modeling methods do not make full use of the physical knowledge in the engineering system. 
Some works such as \cite{chakraborty2021transfer} build a multi-fidelity model on the base of Physics Informed Neural Network (PINN), which can make full use of prior physical knowledge to train the model without labeled low-fidelity data. However, these methods are mainly applied to solve PDE equations and cannot be used for surrogate modeling.
Considering the task at hand, summarize the problems remain when jointing the multi-fidelity modeling with CNN for temperature field prediction: 

\begin{itemize}
	\item General deep multi-fidelity modeling methods explore the relationship between low and high fidelity data and require multiple networks in series to achieve the multi-fidelity model, leading to structure redundancy and error accumulation.
	
	\item Existing multi-fidelity modeling methods adopt supervised learning, relying on a large amount of labeled low-fidelity data. Although acquisition costs are low, a mass of low-fidelity data still increases the cost of surrogate model construction.
\end{itemize} 

In order to solve the first problem, this paper proposes a deep multi-fidelity model (DMFM) based on the pre-train fine-tune paradigm, implicitly characterizing the multi-fidelity relationship by transferring the pre-trained low-fidelity model to fine-tuned high-fidelity model.
The constructed deep surrogate model is one CNN with two different fidelity projection head. The pre-trained low-fidelity model and the fine-tuned high-fidelity model share the same CNN parameters, which greatly reduces the complexity of the multi-fidelity model and simplifies the model structure.
For the latter problem, this work proposes utilizing self-supervised learning to construct a physical-driven deep multi-fidelity model (PD-DMFM) based on the pre-train fine-tune paradigm.
The self-supervised learning takes advantage of the prior physical knowledge of the temperature field prediction problem, training the low-fidelity model without relying on the labeled low-fidelity data. Therefore, the proposed method can further reduce the amount of data required for training on the basis of multi-fidelity modeling.

To be concluded, the innovations of this paper are as follows:

\begin{enumerate}
	\item This paper proposes employing the pre-train fine-tune paradigm to learn the deep multi-fidelity model (DMFM) with a simplified structure, which significantly reduces the complexity of the multi-fidelity model.
	
	\item This paper proposes a novel self-supervised learning method for the pre-train stage. The trained multi-fidelity model, namely physics-driven deep multi-fidelity model (PD-DMFM), fully utilizes the physics characteristics of the temperature field and can be learned without the labeled low-fidelity data.
	
	\item The performance of proposed DMFM and PD-DMFM is tested through two temperature field prediction problems. The experiments show that both DMFM and PD-DMFM maintain high precision while reducing the number of high-fidelity data required. The accuracy of DMFM is relatively higher than PD-DMFM under the same scale of high-fidelity data. While PD-DMFM further reduces the cost of obtaining low-fidelity data and achieves competitive performance. 
\end{enumerate}

This paper is structured as follows. The temperature field prediction problem is described in Sec.~\ref{sec:2}. Sec.~\ref{sec:3} presents the general framework of the deep multi-fidelity model for temperature field prediction. The experiments in Sec.~\ref{sec:exp} show the DMFM and PD-DMFM test performance in two cases with various difficulties. Finally, the main conclusions of this paper are stated in Sec.~\ref{sec:conclusions}.

\section{Problem statement}
\label{sec:2}
In this section, the temperature field prediction problem is introduced, which is similar to the problem proposed in \cite{chen2016optimization,chen2017heat,aslan2018heat}. Consider placing a number of electronic components that emit heat at work in a square container with a heat dissipation hole, and heat can only be dissipated to the outside through the heat dissipation hole. In order to simplify the problem, it is considered that only heat conduction occurs in the container without heat convection and radiation. 

\begin{figure}[htbp]
	\centering 
	\includegraphics[width=0.4\textwidth]{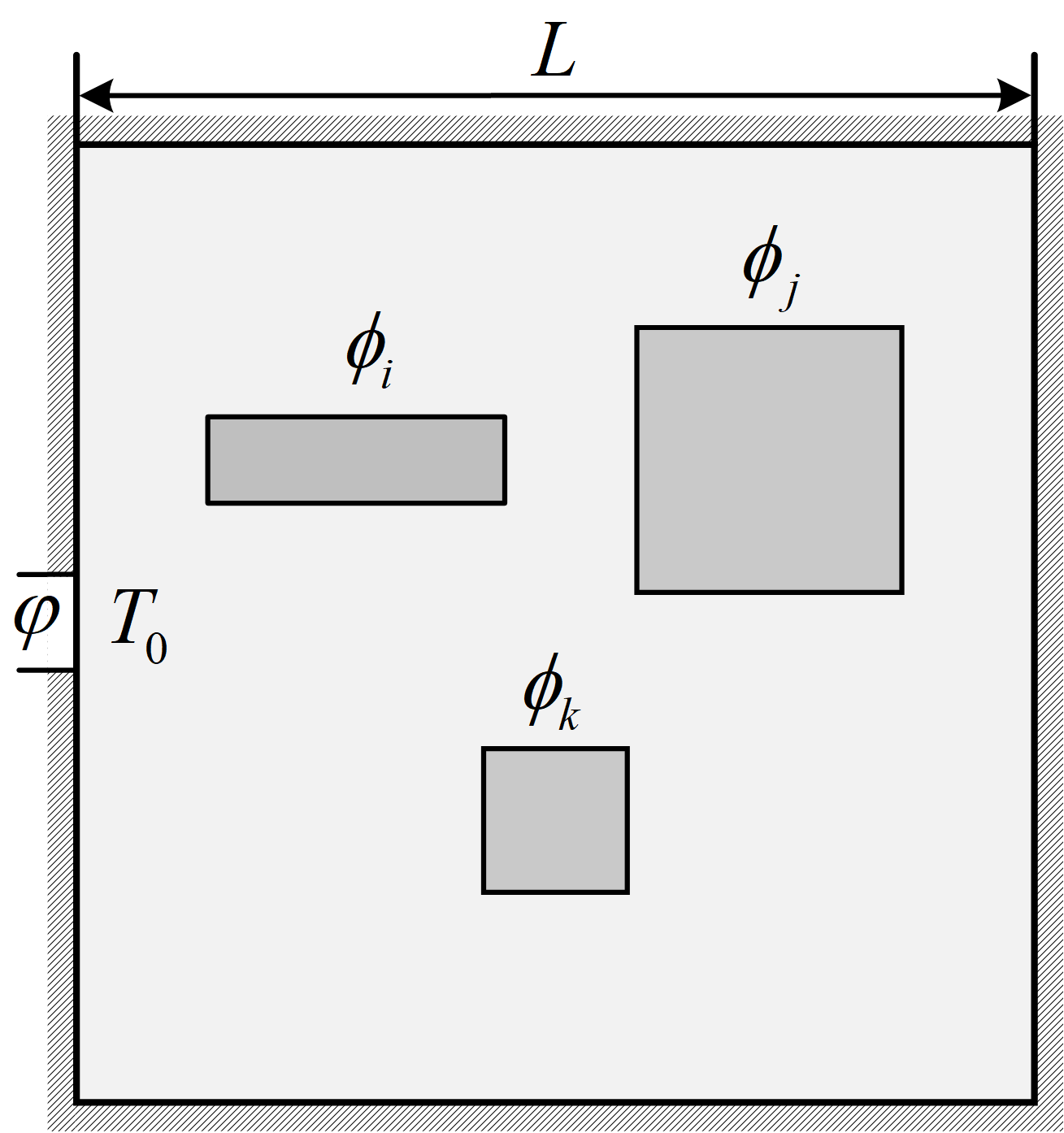} 
	\caption{Two-dimensional volume-to-point problem}  
	\label{fig:1}
\end{figure}

The purpose of temperature field prediction task is to predict the steady temperature inside the container during the work of electronic components. This problem can be modeled as a two-dimensional volume-to-point (VP) problem as shown in Fig.~\ref{fig:1}. The two-dimensional square region is surrounded by an adiabatic boundary of length $L$, there are a heat dissipation hole of length $\delta$ at the midpoint of the left boundary, and the temperature at the hole remains constant. A number of $N$ electronic components, be seen as heat sources, are placed in the two-dimensional region. The components are modeled as the shape of square or rectangle with different lengths and widths.

In the two-dimensional region, the steady temperature field with heat sources can be calculated by solving Poisson equation with Dirichlet and Neumann boundary conditions, which is formulated as
\begin{equation}
	\begin{aligned}
		\frac{\partial}{\partial x}(k\frac{\partial T}{\partial x})+\frac{\partial}{\partial y}(k\frac{\partial T}{\partial y})+\phi(x,y)=0 \\
		{\rm Boundary}: T=T_0\quad {\rm or}\quad k\frac{\partial T}{\partial n}=0 
	\end{aligned}
	\label{equ:1}
\end{equation}
where $T$ is the temperature of layout area, $k$ is the thermal conductivity of layout area and $\phi(x,y)$ is the intensity distribution function of heat sources determined by the location of the heat source, which can be expressed as
\begin{equation}
	\phi(x,y)=
	\begin{cases}
		\phi_i, & (x,y)\in\Gamma \\
		0, & (x,y)\in\Gamma
	\end{cases}
	\label{equ:2}
\end{equation}
where $\phi_i$ represents the intensity of $i$-th heat source and $\Gamma$ denotes the layout area covered by the heat source. The change of the position of the heat source can affect the intensity distribution function and eventually change the steady temperature field in the layout area.
In the boundary conditions of the governing Eq.~\ref{equ:1}, $T_0$ is the temperature of constant temperature boundary, $h$ is the convective heat transfer coefficient. 

In this paper, the temperature field prediction problem is considered as an image-to-image regression task, the layout is taken as the input of the model, and the corresponding temperature field is taken as the output of the model. To solve this dense presiction problem, the convolutional neural network (CNN) is employed to construct our deep multi-fidelity model following the previous works \cite{song2021transfer,chen2020heat,chen2021deep}. CNN has strong nonlinear fitting ability and is suitable in solving high-dimensional prediction problems. The typical CNN has multiple layers, and the output features in the $l$-th layer of CNN can be expressed by the following formula:
\begin{equation}
	{{X}^{l}}=pool\left( {{A}^{l}}\left( {{W}^{l}}{{X}^{l-1}}+{{b}^{l}} \right) \right)
\end{equation}
where ${X}^{l}$ and ${X}^{l-1}$ respectively represents the feature of $l$-th layer and $(l-1)$-th layer, ${W}^{l}$ and ${b}^{l}$ respectively represents the kernel weight and bias term of $l$-th layer, ${A}^{l}$ is the activation function of $l$-th layer, and $pool(\cdot)$ is the pooling function. Here we use $\theta$ to denote the kernel weight ${W}^{l}$ and bias term ${b}^{l}$. In practical using, the optimum parameters $\theta$ of CNN need to be determined by minimizing an loss function according to empirical risk minimization \cite{vapnik1992principles}.

\section{Proposed Method}
\label{sec:3}
Following sections detail the proposed deep multi-fidelity modeling method for the temperature field prediction. 
The deep multi-fidelity model (DMFM) based on pre-train and fine-tune paradigm is firstly presented in Sec.~\ref{sec:3.1}, following the introduction for the architecture design of DMFM in Sec.~\ref{sec:3.2}. Finally, the physics-driven deep multi-fidelity model (PD-DMFM) with self-supervised learning is proposed in Sec.~\ref{sec:3.3} on the base of pre-train and fine-tune paradigm.

\subsection{Deep Multi-Fidelity Model based on Pre-train and Fine-tune Paradigm}\label{sec:3.1}
The multi-fidelity modeling aims to generalize from a combination set of limited high-fidelity data $D^{h}=\{(x^{h}_i,y^{h}_i)\}^N_{i=1}$ and abundant low-fidelity data $D^l=\{(x^l_i,y^l_i)\}^M_{i=1}$, where $M\gg N$. Most of previous the multi-fidelity modeling methods explore the relationship between low- and high-fidelity data to construct the model. Specifically, a low-fidelity model is firstly trained on the low-fidelity data, and another model is trained on $\{(y^{l}_i,y^{h}_i)\}^N_{i=1}$ to learn the mapping from the low-fidelity data to corresponding high-fidelity data. The accuracy of model describing the relationship between low- and high-fidelity data affects the performance of multi-fidelity model. However, the data pairs $(y^{l}_i,y^{h}_i)$ used to represent the relationship are limited to the number of high-fidelity data, which futher restricts the characterization ability of multi-fidelity model. Besides, during the process of subsequent design optimization or uncertainty quantification, two models constructed by the above method need to be used jointly, that is structurally redundant. 

\begin{figure}[htbp]  
	\centering  
	\includegraphics[width=\textwidth]{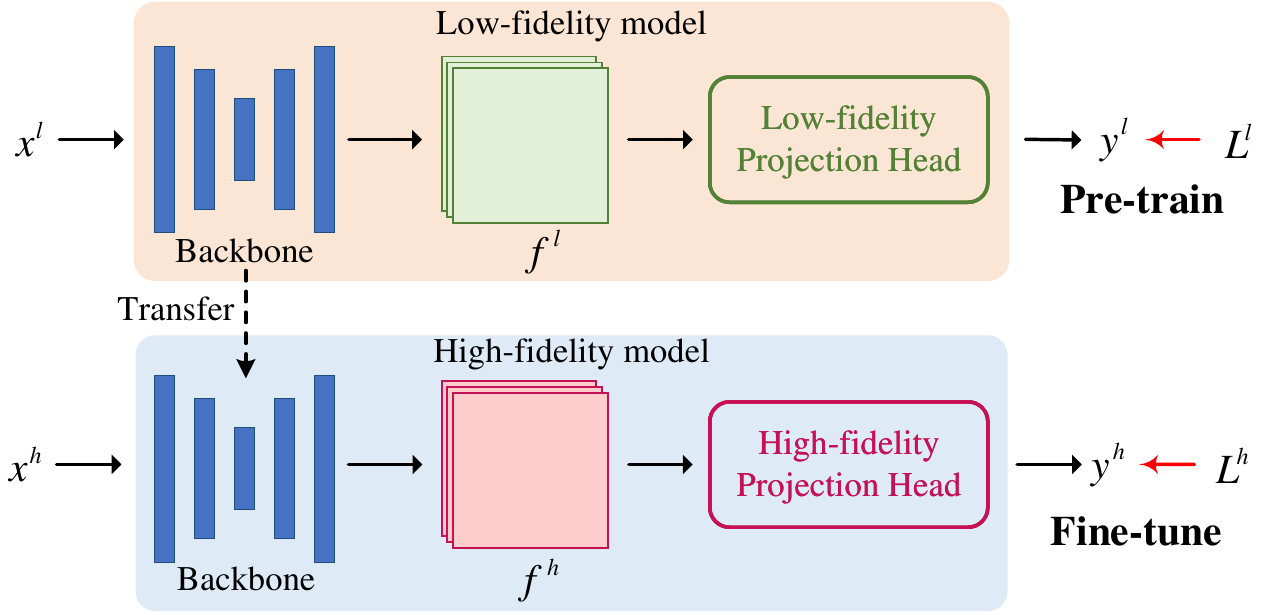}  
	\caption{The framwork of deep multi-fidelity model with pre-train and fine-tune paradigm}  
	\label{fig:2}
\end{figure}

Instead of directly mapping low-fidelity data to high-fidelity data, we construct the multi-fidelity model utilizing the pre-train and fine-tune paradigm, implicitly characterize the multi-fidelity relationship by the aid of the model portability. The framwork of proposed deep multi-fidelity modeling method in this paper is shown in Fig.~\ref{fig:2}.

The deep multi-fidelity model mainly includes two parts, namely backbone and multi-fidelity projection head. The role of backbone is to extract features from the multi-fidelity input data. In order to map the extracted features to the output, a projection head is connected behind the backbone. According to the different fidelity of output data, the projection head can be divided into low-fidelity projection head and high-fidelity projection head. 

\begin{algorithm}
	\renewcommand{\algorithmicrequire}{\textbf{Input:}}
	\renewcommand{\algorithmicensure}{\textbf{Output:}}
	\caption{Deep multi-fidelity model with pre-train and fine-tune paradigm}
	\label{alg1}
	\begin{algorithmic}[1]
		\REQUIRE {Low-fidelity dataset $D^{l}=\{(x^{l}_i,y^{l}_i)\}^M_{i=1}$; High-fidelity dataset $D^{h}=\{(x^{h}_i,y^{h}_i)\}^N_{i=1}$,}
		\ENSURE Backbone $\mathcal B$; Low-fidelity projection head ${\mathcal H}^l$; High-fidelity projection head ${\mathcal H}^h$\\
		\emph{\#Pre-train}
		\FOR {$minibatch \{(x^{l}_k,y^{l}_k)\}^B_{k=1} \in D^l$}  
		\FOR {$k\in\{1,...,B\}$}
		\STATE $\hat{y}^l_k={\mathcal H}^l({\mathcal B}(x^l_k))$;
		\ENDFOR
		\STATE Update $H^l$ and ${\mathcal B}$ to minimize $L^l$ of $\{(\hat{y}^l_k,y^{l}_k)\}^B_{k=1}$;
		\ENDFOR \\
		\emph{\#Fine-tune}
		\FOR {$minibatch \{(x^{h}_k,y^{h}_k)\}^B_{k=1} \in D^h$}
		\FOR {$k\in\{1,...,B\}$}
		\STATE $\hat{y}^h_k={\mathcal H}^h({\mathcal B}(x^h_k))$;
		\ENDFOR
		\STATE Update ${\mathcal H}^h$ and ${\mathcal B}$ to minimize $L^h$ of $\{(\hat{y}^h_k,y^{h}_k)\}^B_{k=1}$;
		\ENDFOR
		\RETURN Deep multi-fidelity model ${\mathcal H}^h({\mathcal B}(*))$.
	\end{algorithmic}  
\end{algorithm}

Specifically, the constructing steps for deep multi-fidelity model with pre-train and fine-tune paradigm is provided below:
\paragraph{Pre-train} Pre-train the model on the abundant low-fidelity data. Firstly, initialize the backbone ${\mathcal B}$ and the low-fidelity projection head ${\mathcal H}^l$. Given a low-fidelity input $x^{l}$, the backbone extracts low-fidelity feature $f^{l}={\mathcal B}(x^l)$, and the low-fidelity projection head predicts $\hat{y}^l={\mathcal H}^l(f^{l})$. Then the prediction $\hat{y}^l$ is supervised by its corresponding ground-truth ${y}^l$ to train the backbone ${\mathcal B}$ and the low-fidelity projection head ${\mathcal H}^l$. As the temperature field prediction problem is defined as an image regression task, the mean absolute loss function is employed as the constraint which is defined as:
\begin{equation}
	L^{l}=\frac{1}{\lvert D^{l}\rvert}\sum_{x^{l}\in D^{l}} \frac{1}{H^l\times W^l}\sum_{i=0}^{H^l\times W^l} \lvert y^l_i-\hat{y}^l_i \rvert,
\end{equation}
where $W^l$ and $H^l$ represent the width and height of low-fidelity input data. Through the pre-train, the backbone can learn rough mapping from input to output on the low-fidelity data. The rough mapping can be used as a basis, making it possible to quickly learn accurate mapping from limited high-fidelity data. 
\paragraph{Fine-tune} Fine-tune the model on the limited high-fidelity data. We firstly transfer the pre-trained ${\mathcal B}$ as the backbone of high-fidelity model, and replace the low-fidelity projection head ${\mathcal H}^l$ as the high-fidelity projection head ${\mathcal H}^h$ to matche the dimension of high-fidelity data. Then, initialize the parameters of the high-fidelity mapping head ${\mathcal H}^h$, and retrain the backbone ${\mathcal B}$ on the high-fidelity data $D^{h}$. Given a high-fidelity input $x^h$, the backbone extracts high-fidelity feature $f^{h}={\mathcal B}(x^h)$, and the high-fidelity projection head predicts $\hat{y}^h={\mathcal H}^h(f^{h})$. Simalar as pre-train stage, the prediction $\hat{y}^h$ is supervised by its corresponding ground-truth ${y}^h$ to train ${\mathcal B}$ and ${\mathcal H}^h$. The constraint can be formulated as:
\begin{equation}
	L^{h}=\frac{1}{\lvert D^{h}\rvert}\sum_{x^{h}\in D^{h}} \frac{1}{H^h\times W^h}\sum_{i=0}^{H^h\times W^h} \lvert y^h_i-\hat{y}^h_i \rvert,
\end{equation}
where $W^h$ and $H^h$ represent the width and height of high-fidelity input data. After training, we can get the deep multi-fidelity model $\mathcal M$, where $\mathcal M(*)={\mathcal H}^h({\mathcal B}(*))$.

\subsection{Architecture Design of Deep Multi-Fidelity Model}
\label{sec:3.2}
\begin{figure}[htbp] 
	\centering  
	\includegraphics[width=\textwidth]{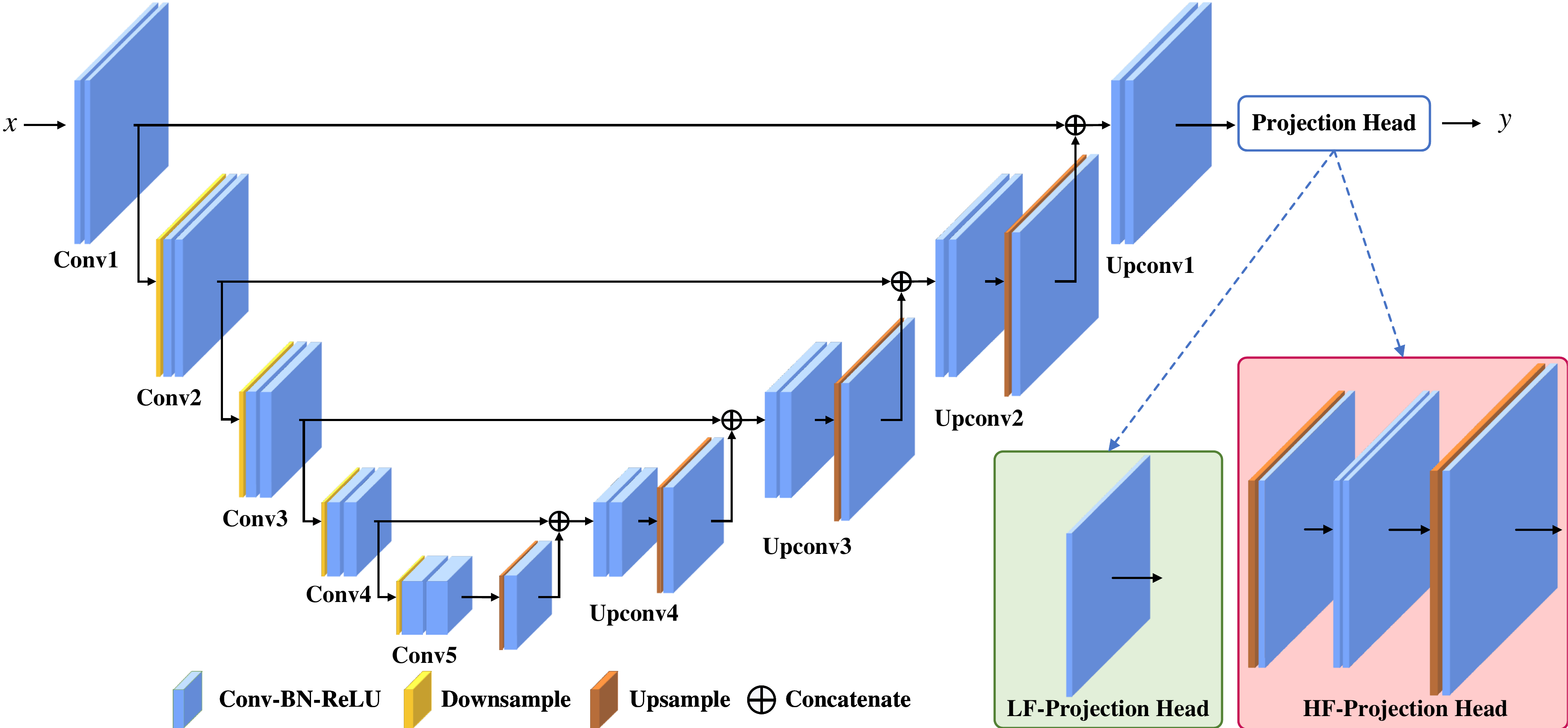}  
	\caption{Architecture of deep multi-fidelity model}  
	\label{fig:3}
\end{figure}

The proposed deep multi-fidelity model consists of the backbone and multi-fidelity projection head, whose architecture is shown in Fig.~\ref{fig:3}. 

\paragraph{Backbone} In order to better extract features from the input data, U-net is adopted as the backbone of model.  
U-net is an effective CNN structure for image-to-image regression, which has a wide range of applications in medical image segmentation and other issues. It can well capture the overall and detailed characteristics of the image, and has the advantages of multi-scale fusion and processing large images.  

The backbone consists of encoder and decoder: 

The encoder part is the image feature extraction part, which is composed of five convolution blocks. Each block contains two convolution layers and the corresponding batch normalization and Relu function activation operations. With the deepening of the network, the channel dimension of the image feature map, namely the depth of the feature map, gradually increases, and the receptive field of each feature map also gradually increases. The last four blocks of the encoder are connected to a down-sampling layer before entering the convolution layer, whose role is to reduce the size of the image feature map to avoid the excessive parameters of the feature map and reduce the amount of network calculation.  

The decoder part is the up-sampling part, which gradually sample the feature map to the size of the model  output. The decoder is composed of four up-sampling convolution blocks, each of which is also composed of two convolution layers, corresponding batch normalization and Relu activation functions. Before entering each block module, the feature outputting from the previous layer module is up-sampled and convolved, and the channel dimension skip layer splicing operation is performed with the same size feature map corresponding to the encoder part.  The purpose of the sampling operation is to gradually decode the feature map which has been deeply fused into the output image size.  Skip layer splicing operation can fuse the shallow features of the network into the image decoding part, which can fully retain the image details and make the output image more refined.   

\paragraph{Multi-fidelity projection head} The low-fidelity projection head consist of a convolution layer, whose main purpose is mapping the feature map output by the backbone to the low-fidelity temperature field image. While the purpose of the high-fidelity projection head is mapping the output feature map of the backbone to the high-fidelity temperature field image. Since the size of the low-fidelity temperature field image is different from that of the high-fidelity temperature field image, the high-fidelity projection head contains two up-sampling layers and a block module composed of two convolution layers, batch normalization and relu activation function, so as to map the feature with smaller size to the temperature field image with larger size. The specific structure of the projection head is shown in Fig.~\ref{fig:3}
\subsection{Physics-Driven Self-Supervised Learning for the Pre-train stage}
\label{sec:3.3}  
The multi-fidelity data is usually obtained through numerical calculations or physical experiments. Although the cost of label acquisition for low-fidelity data is cheaper than high-fidelity data, getting a mass of labels for training the low-fidelity model is still cumbersome. To further reduce the cost of labels, the self-supervised learning is proposed for the pre-train of deep multi-fidelity model.

Different from the supervised learning, the self-supervised learning mine the information of the data itself for training, without relying on the label. The key to applying self-supervised learning to temperature field prediction problem is how to utilize prior knowledge. Physics informed neural networks (PINN) \cite{raissi2019physics} is a method of using prior knowledge successfully, address the over-reliance of the labels in supervised learning. PINN encode physical conservation laws and prior physical knowledge into the model by minimizing the loss function which is constructed by governing equation (ODE / PDE) of the problem. The output of model satisfies the governing equation with the training, so as to complete the solution of the ODE / PDE. 
Inspired by PINN, we use the heat conduction equation and the finite difference method to construct the loss function for the pre-train, realizing the physics-driven self-supervised learning. Concretely, transform the heat conduction equation shown in Eq.~\ref{equ:1} into a difference equation as follows:
\begin{equation}
	\begin{aligned}
		\boldsymbol{\phi}(x,y) & +  \frac{\boldsymbol{T}\left(x_{i+1},{y_j} \right)-2\boldsymbol{T}\left( {x_i},{y_j} \right) + \boldsymbol{T}({x_{i-1}},y_j)}{\Delta{x}^{2}}k \\& + 
		\frac{\boldsymbol{T}\left( {{x}_{i}},{{y}_{j+1}} \right)-2\boldsymbol{T}\left( {{x}_{i}},{{y}_{j}} \right)+ \boldsymbol{T}({{x}_{i}},{{y}_{j-1}})}{{\Delta{y}^{2}}}k =0
	\end{aligned}
	\label{equ:6}
\end{equation}
where ${x}_{i}$ and ${y}_{j}$ denote \emph{i}-th and \emph{j}-th mesh point  respectively, $T(x_i,y_j)$ is the temperature at $(x_i,y_j)$, $\Delta{x}$ and $\Delta{y}$ represent the mesh size in \emph{X} and \emph{Y} directions respectively.
Since the multi-fidelity data used in this paper is partitioned by square mesh as shown in the Fig.~\ref{fig:3}, the mesh size $\Delta{x}$ in \emph{X} direction and $\Delta{y}$ in \emph{Y} direction is equal. 
The Eq.~\ref{equ:6} can be converted to the following form,  
\begin{equation}
	\begin{aligned}
	\boldsymbol{T}\left( {{x}_{i}},{{y}_{j}} \right)= & \frac{1}{4}(\Delta {{x}^{2}}\boldsymbol{\phi} (x,y)+\boldsymbol{T}\left( {{x}_{i+1}},{{y}_{j}} \right)+\boldsymbol{T}({{x}_{i-1}},{{y}_{j}}) \\& +\boldsymbol{T}\left( {{x}_{i}},{{y}_{j+1}} \right)+\boldsymbol{T}({{x}_{i}},{{y}_{j-1}}))
	\end{aligned}
	\label{equ.7}
\end{equation}

Ideally, the output of the surrogate model, that is the temperature at $(x_i,y_j)$, should satisfy Eq.~\ref{equ.7}. 
Thus the physical-driven loss function for low-fidelity data is founded, which is shown as follows:
\begin{equation}
	L^{l}=\frac{1}{\lvert D^{l}\rvert}\sum_{x^{l}\in D^{l}} \frac{1}{H^l\times W^l}\sum_{i=0}^{H^l\times W^l} \lvert \boldsymbol{T}_i-\hat{y}^l_i \rvert,
\end{equation}

By minimizing the physical-driven loss function, the output of model, that is the temperature field satisfies the heat conduction difference equation. It is worth mentioning that we only emply the physical-driven loss function for the pre-train stage. During the fine-tune stage, considering the situation that high-fidelity data can be obtained through physical experiments, the physical-driven training method is not able to adopt since the heat conduction equation cannot accurately describe the actual physical phenomena.

\section{Experiment}
\label{sec:exp}
\subsection{Data preparation}
\label{sec:4.1}
To fully evaluate the performance of the deep multi-fidelity model for temperature field prediction, this paper introduces two layout situations according to the different shapes of internal components, namely simple and complex layout, in which the temperature prediction of complex layout is more difficult. For both simple and complex layout, the adiabatic boundary length $L$ is set to $0.1m$, the thermal conductivity $k$ of layout area is $1 w/(m*K)$, and the temperature at the heat dissipation hole is fixed as $298K$.
The simple layout has $20$ square electronic components with the same size of $0.01m \times 0.01m$, the heat source strength of each component is $10000 w/m^2$.
For the complex layout, there are $12$ electronic components that are rectangles with different lengths. The heat source strength and shape size of all components are shown in Table.~\ref{tab:1}.

\begin{table}[htbp]
	\centering
	\caption{Complex layout components}
	\begin{tabular}{ccc}
		\toprule
		component number & size$(m \times m)$ & heat source intensity $(w/m^2)$ \\
		\midrule
		1 & 0.016 $\times$ 0.012 & 4000 \\
		2 & 0.012 $\times$ 0.006 & 16000 \\
		3 & 0.018 $\times$ 0.009 & 6000 \\
		4 & 0.018 $\times$ 0.012 & 8000 \\
		5 & 0.018 $\times$ 0.018 & 10000 \\
		6 & 0.012 $\times$ 0.012 & 14000 \\
		7 & 0.018 $\times$ 0.006 & 16000 \\
		8 & 0.009 $\times$ 0.009 & 20000 \\
		9 & 0.006 $\times$ 0.024 & 8000 \\
		10 & 0.006 $\times$ 0.012 & 16000 \\
		11 & 0.012 $\times$ 0.024 & 10000 \\
		12 & 0.024 $\times$ 0.024 & 20000 \\
		\bottomrule
	\end{tabular}
\label{tab:1}
\end{table}

\begin{figure}
	\centering 
	\includegraphics[width=0.7\textwidth]{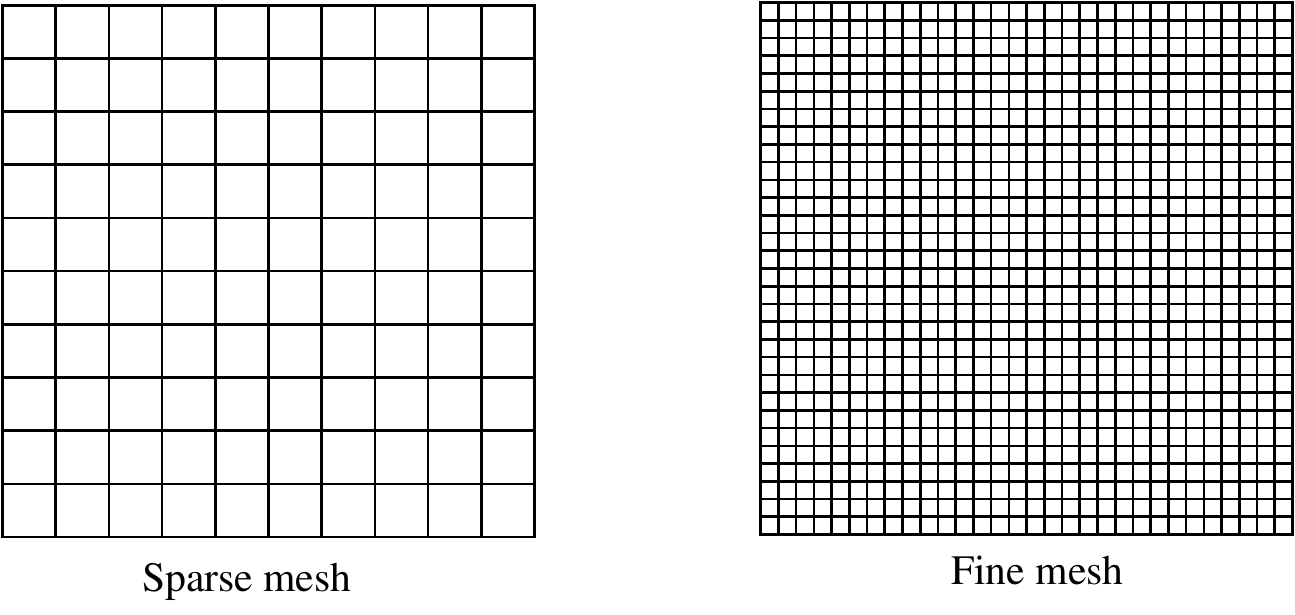}
	\caption{Low and high fidelity data meshing}  
	\label{fig:4}
\end{figure}

The multi-fidelity data used in this paper are obtained through numerical simulation. We distinguish between low and high fidelity data through the mesh density, where low-fidelity data sparsely meshes and high-fidelity data finely meshes, as shown in Fig.~\ref{fig:4}. 
Specifically, the low-fidelity temperature field is divided into $50\times50$ grids and the high-fidelity temperature field into $200\times200$ grids.
Then the temperature field of the corresponding layout is calculated by the finite difference method (FDM) and used as the label of training samples. 

For both simple and complex layout, this paper prepares $2000$ labeled low-fidelity data for the training of the DMFM during the pre-train stage. Since the self-supervised learning does not rely on labels, $9000$ low-fidelity data without labels, whose the acquisition is almost free, are generated for the training of PD-DMFM during the pre-train stage.
In order to explore the performance variation with change in the number of high-fidelity data, this paper uses different amounts of high-fidelity data selected from $\{10, 20, 30, 40, 50, 100, 200, 500, 1000, 2000\}$ to train DMFM and PD-DMFM during the fine-tune stage.

\subsection{Implementation details}
\label{sec:4.2}
The proposed DMFM and PD-DMFM are implemented based on the Pytorch framework in all experiments. The model training is completed on a high-performance computer server, and its computing resource allocation is Intel(R) Xeon(R) Gold 6242 CPU @ 2.80 GHz, Nvidia GTX 3090 GPU with 24GB VRAM, and 500 GB RAM. 

This paper applies Kaiming initialization \cite{he2015delving} to initialize the parameters of the backbone, low and high-fidelity projection head. 
Ranger optimizer \cite{Ranger}, a synergistic optimizer combining RAdam (Rectified Adam) \cite{liu2019variance} and LookAhead \cite{zhang2019lookahead}, is adopted to train our DMFM and PD-DMFM. In order to ensure the fairness, the parameters of the optimizer used in experiments are consistent. The initial learning rate of the backbone is $\eta=0.01$ and $\eta=0.001$ for the pre-train and fine-tune stages, respectively. For both low and high-fidelity projection head, the initial learning rate is set to $0.01$. Besides, Cosine Annealing Warm Restarts scheduler \cite{loshchilov2016sgdr} is selected as our learning rate policy where the learning rate varies according to the following equation:
\begin{equation}
	\eta_t=\eta_{min}+\frac{1}{2}(\eta_{max}-\eta_{min})(1+cos(\frac{T_{cur}}{T_i}\pi))
\end{equation}
where $\eta_{min}$ is the minimum learning rate which is set to $1e^{-7}$ in our experiments, $\eta_{max}$ is the maximum learning rate (the initial learning rate is employed as the maximum learning rate), $T_{cur}$ is the number of epochs since the last restart, and $T_{i}$ is the number of epochs between two restarts which can be expressed as:
\begin{equation}
	T_i=
	\begin{cases}
		T_0, & i=0 \\
		T_{i-1}\times T_{mult}, &  i>0
	\end{cases}
\end{equation}
where $T_0$ is the number of epochs for the initial restart which is set to $10$, $T_{mult}$ is the factor of the restart which is set to $2$ in this paper. For further details about the scheduler and optimizer, please referring the official documentation for Pytorch \cite{paszke2019pytorch}.

In all experiments, the epoch is fixed as 150, the batch size during the pre-train stage is set to 16.
During the fine-tune stage, a large batch size can affect the model's performance due to the small amount of data. Thus this paper sets the batch size of experiments using 10 to 200 high-fidelity data to 1. 
For a fair comparison, the batch size of experiments with 500, 1000, and 2000 high-fidelity samples are set to 2, 4, and 8, respectively, ensuring the number of iterations is consistent as much as possible.

\subsection{Evaluation Metrics}
\label{sec:4.3}
In order to evaluate the performance of proposed DMFM, we select the following evaluation metrics according to the setting in \cite{chen2021deep}: 

\textbf{The Mean Absolute Error (MAE)}. The MAE of the prediction on the whole temperature field is an important metric to evaluate the predictive ability of surrogate models, which is expressed as:
\begin{equation}
	\text{MAE}=\frac{1}{N^2}\sum_{i=1}^N\sum_{j=1}^N\left\vert \hat{Y}_{ij}-Y_{ij} \right\vert
\end{equation}
where $N$ is the mesh numbers on one side of temperature field, and set to 50 and 200 for low and high-fidelity data, respectively.

\textbf{Component-constrained MAE (CMAE)}. 
CMAE is the mean absolute error of the prediction on the field that components cover. The temperature of the domain where components cover is usually our concern. Consider using mask $M$ to represent the area covered by components. The value of 1 in the mask denotes the component covering, and 0 indicates non-component coverin
CMAE can be computed as:
\begin{equation}
	\text{CMAE} =\frac{1}{M}\sum{(M\otimes \left| \hat{Y}-Y \right|)}
\end{equation}
where $M$ is the mask of components, $\otimes$ represents the elementwise multiplication. 

\textbf{The Absolute Error of the Maximum Temperature (MT-AE)}. MT-AE is the absolute error of the maximum temperature. The maximum temperature is another indicator of interest to us, which must be forbidden from exceeding the specified safety threshold. MT-AE can be calculated as below.
\begin{equation}
	\text{MT-AE}=\left| \max (\hat{Y})-\max (Y) \right|
\end{equation}
where $\hat{Y}$ and $Y$ are the prediction and ground-truth of the temperature matrix, respectively.

\subsection{Temperature field prediction performance for simple layout data}

\begin{figure}[htbp]
	\centering 
	\includegraphics[width=\textwidth]{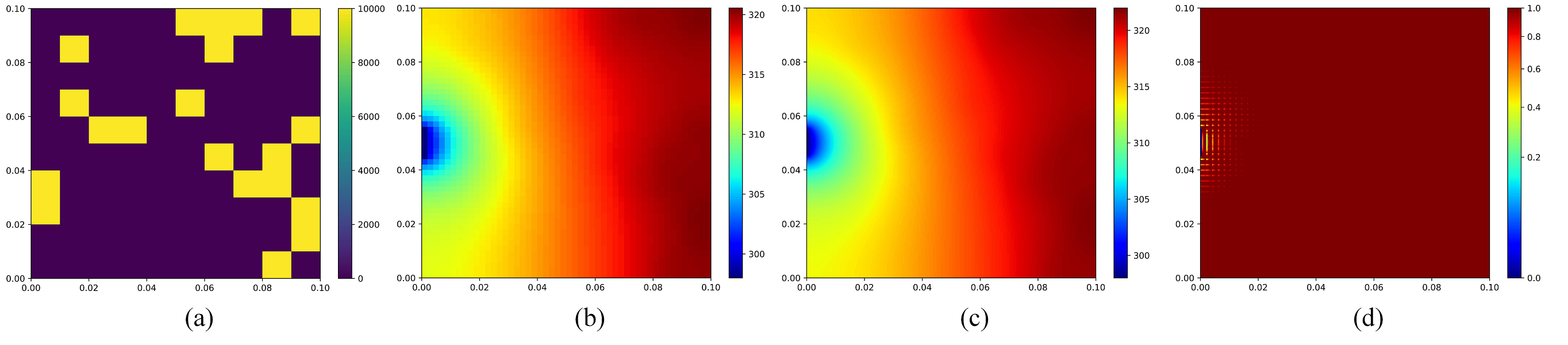}
	\caption{The multi-fidelity data of simple layout. (a) simple layout, (b) the low-fidelity temperature field, (c) the high-fidelity temperature field, (d) the error between the high and the interpolated low fidelity data}  
	\label{fig:5}
\end{figure}

This section considers the simple layout situation. The low and high fidelity data used in the experiments are shown in Fig.~\ref{fig:5}. We can see that the low-fidelity data can reflect the trend of the temperature field but roughly in detail. Fig.~\ref{fig:5}(d) is the error between the high-fidelity data and the interpolated low-fidelity data, the errors of almost all pixels are greater than $1K$. 

The DMFM and PD-DMFM are trained with the low-fidelity date and different amounts of high-fidelity data.  
In addition, the single-fidelity model (SFM) is also trained with only high-fidelity data to compare the performance with the proposed DMFM and PD-DMFM. The parameter settings of the single-fidelity experiment are consistent with the multi-fidelity experiment. Tab.~\ref{tab:2} exhibits the result evaluated by the metrics in Sec.~\ref{sec:4.3}

\begin{figure}[htbp]
	\centering 
	\includegraphics[width=\textwidth]{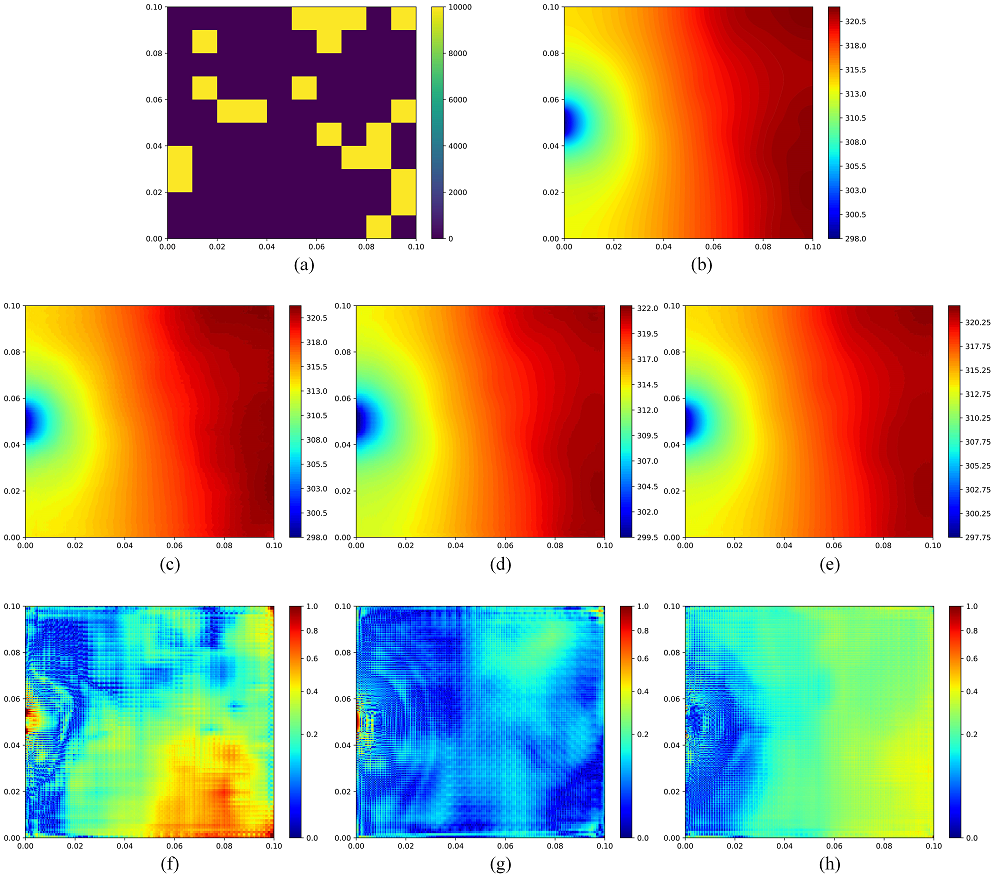}\\
	\caption{Visualization results for the simple layout. (a)simple layout, (b)ground-truth, (c)SFM prediction, (d)DMFM prediction, (e)PD-DMFM prediction, (f)prediction error of SFM, (g)prediction error of DMFM, (h)prediction error of PD-DMFM.}  
	\label{fig:6}
\end{figure}

\paragraph{The comparison of different methods}
The results in Tab.~\ref{tab:2} show that both DMFM and PD-DMFM performance on each metric is much better than SFM under different amounts of high-fidelity data. Taking models trained by $10$ high-fidelity data as an example, the MAE of DMFM and PD-DMFM are $0.2202$ and $0.2149$, $63.5\%$ and $64.3\%$ lower than SFM. For the CMAE, where the value is $0.207$ and $0.1783$, the error reduction separately reaches  $43.5\%$ and $51.4\%$. The performance improvement is even more pronounced on the MT-AE indicator, where the value is $0.4126$ and $0.285$, with an error reduction of $61.7\%$ and $73.6\%$, respectively. As for the situations under other amounts of high-fidelity data, the promotion of DMFMF and PD-DMFM over SFM is also enormous. 
The DMFMF and PD-DMFM show great improvement in relative performance and excellent performance in absolute value. The predictions of SFM, PD-DMFM, and DMFM trained by $50$ high-fidelity samples are visualized in Fig.\ref{fig:6}. One can observe that both PD-DMFM and DMFM accurately predict the trend of the temperature field. Errors of most points in the temperature field predicted by PD-DMFM and DMFM are around $0.2K$.

\begin{table}[htbp]
	\centering
	\caption{Performance comparison of SFM, DMFM and PD-DMFM under different amounts of high-fidelity data for the simple layout.}
	\resizebox{\textwidth}{!}{ 
		\begin{tabular}{ccllllllllll}
			\toprule
			\multicolumn{2}{c}{Amounts of high-fidelity data}&  10    & 20    & 30    & 40    & 50    & 100   & 200   & 500   & 1000  & 2000 \\
			\midrule
			\multirow{3}{*}{SFM} & MAE   & 0.6027 & 0.5879 & 0.3118 & 0.2093 & 0.1652 & 0.1104 & 0.0723 & 0.0338 & 0.0241 & 0.0211 \\
			& CMAE  & 0.3668 & 0.2215 & 0.1897 & 0.1593 & 0.1333 & 0.0794 & 0.0494 & 0.0326 & 0.0295 & 0.0246 \\
			& MT-AE  & 1.0792 & 1.2972 & 0.5866 & 0.5971 & 0.447 & 0.1884 & 0.1202 & 0.0558 & 0.0646 & 0.0387 \\
			\midrule
			\multirow{3}{*}{DMFM} & MAE   & 0.2202 & 0.1955 & \textbf{0.1009} & \textbf{0.0979} & \textbf{0.0938} & \textbf{0.0613} & \textbf{0.042} & \textbf{0.0235} & \textbf{0.0177} & 0.0163 \\
			& CMAE  & 0.207 & 0.1592 & \textbf{0.0916} & 0.0941 & 0.0941 & \textbf{0.0569} & \textbf{0.041} & 0.0289 & 0.022 & 0.0205 \\
			& MT-AE  & 0.4126 & \textbf{0.2727} & \textbf{0.1484} & \textbf{0.1392} & 0.1561 & \textbf{0.0763} & \textbf{0.0623} & 0.0399 & \textbf{0.0297} & 0.0431 \\
			\midrule
			\multirow{3}{*}{PD-DMFM} & MAE   & \textbf{0.2149} & \textbf{0.1919} & 0.1643 & 0.1204 & 0.1065 & 0.0771 & 0.062 & 0.0267 & 0.0199 & \textbf{0.0149} \\
			& CMAE  & \textbf{0.1783} & \textbf{0.1443} & 0.1229 & \textbf{0.0881} & \textbf{0.0743} & 0.0614 & 0.0454 & \textbf{0.0271} & \textbf{0.0204} & \textbf{0.0151} \\
			& MT-AE  & \textbf{0.285} & 0.2825 & 0.2422 & 0.1868 & \textbf{0.1542} & 0.1147 & 0.0956 & \textbf{0.0395} & 0.0369 & \textbf{0.0286} \\
			\bottomrule
	\end{tabular}}
	\label{tab:2}
\end{table}

Compared with DMFM, PD-DMFM does not rely on labeled low-fidelity data but has lower performance. The results in Tab.~\ref{tab:2} denoted that DMFM performs better than PD-DMFM in most metrics under different amounts of high-fidelity data except 10, 20, and 2000. 
When the high-fidelity data size is $30$, the gap on MAE between DMFM and PD-DMFM is $0.0634$, which is the maximum difference in all situations.
When the high-fidelity data is 10, 20, and 2000, the performance of PD-DMFM is higher than DMFM, but the improvement is small.

\begin{figure}[htbp] 
	\centering 
	\includegraphics[width=0.6\textwidth]{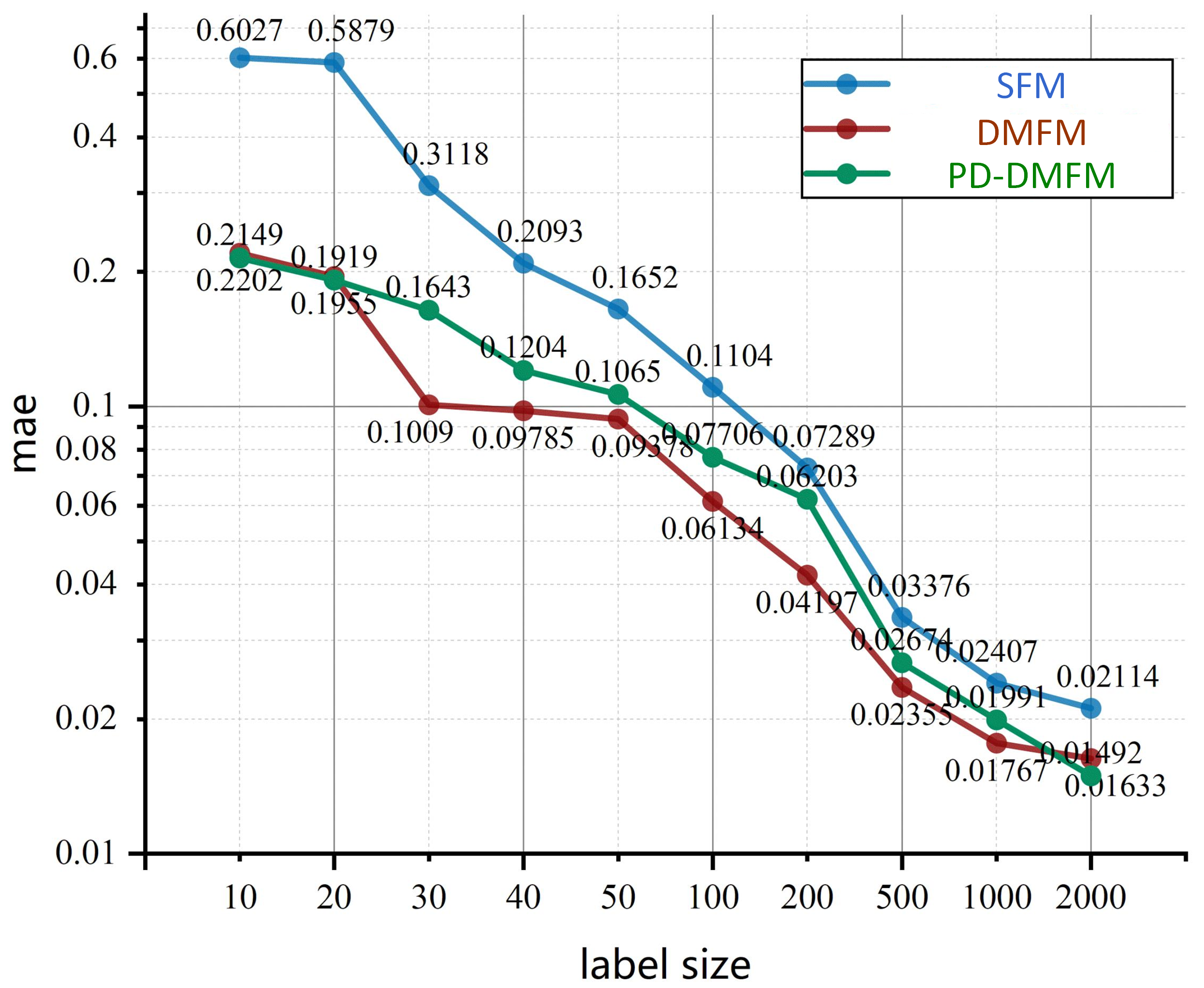}
	\caption{The changes of MAE with different amounts of high-fidelity data.}  
	\label{fig:7}
\end{figure}

\paragraph{The comparison of different high-fidelity data sizes}
The amount of high-fidelity data is an important variable affecting the DMFM and PD-DMFM performance. 
The indicators have improved significantly with the increase in the number of high-fidelity data. The variation of MAE for temperature field prediction with different amounts of high-fidelity data is exhibited in Fig.~\ref{fig:7}. The results denote that the performance increase of DMFM and PD-DMFM relative to SFM is more obvious when the data size is small, and the promotion decreases with the growth of data amount.
When the amount of high-fidelity data is ${10,20,30,40}$, the performance increase of DMFM is more than $50\%$ compared with SFM, while the performance increase of PD-DMFM is more than $40\%$.
The proposed DMFM and PD-DMFM can significantly reduce the requirement of high-fidelity data used for the training model. The MAE of DMFM trained with $30$ high-fidelity samples is $0.1009$, while the MAE of SFM trained with $100$ samples is $0.1104$. In other words, DMFM achieves the same performance as SFM with one-third of the data, heavily reducing the requirement for high-fidelity data. As for PD-DMFM, the MAE is $0.1065$ when the high-fidelity data size is $50$, which is lower than the MAE of SFM using $100$ samples. In conclusion, DMFM and PD-DMFM can achieve the same accuracy as SFM with fewer samples, reducing the demand for the labeled data.

\subsection{Temperature field prediction performance for complex layout data}
\begin{figure}[htbp]
	\centering 
	\includegraphics[width=\textwidth]{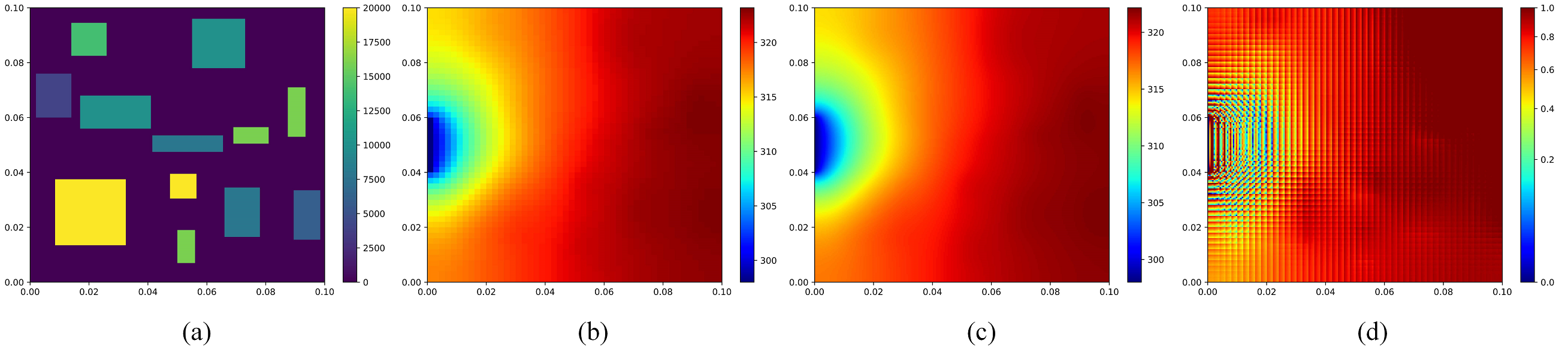}
	\caption{The multi-fidelity data of complex layout. (a) complex layout, (b) the low-fidelity temperature field, (c) the high-fidelity temperature field, (d) the error between the high and the interpolated low fidelity data}  
	\label{fig:8}
\end{figure}

This section considers the complex layout situation. The low and high fidelity data used in the experiments are shown in Fig.~\ref{fig:8}. The rough grid partition leads to an inconsistent representation in the low and high-fidelity data.
Same as the simple layout experiments, DMFM and PD-DMFM are trained with the low-fidelity data and different amounts of high-fidelity data. Besides, SFM is also trained aim of comparing with DMFM and PD-DMFM. The results are shown in Tab.~\ref{tab:3}.

Compared with the simple layout, it is more difficult to construct an accurate mapping from the complex layout to the temperature field due to the inconsistency of component shape. Therefore, the performance of the surrogate model for the complex layout is poor, as shown in the Tab.~\ref{tab:3}. For example, the MAE, CMAE, and MT-AE of SFM trained by 10 high-fidelity data are 0.972, 0.592, and 5.1394, respectively. The surrogate model with such unsatisfactory performance is difficult to apply to practical engineering.
However, benefiting from exploring low-fidelity data, the performance of the proposed DMFM and PD-DMFM is improved with the same amounts of high-fidelity data.
The MAE of DMFM and PD-DMFM are 0.3491 and 0.265, which are $64.1\%$ and $72.7\%$ lower than SHF. For the CMAE, where the value is 0.2514 and 0.1771, the error reduction separately reaches $57.5\%$ and $70.1\%$. For the MT-AE, where the value is 0.7241 and 0.4363, the error reduction is $85.9\%$ and $91.5\%$, respectively. The results show that the proposed DMFM and PD-DMFM still perform well on the difficult problem. 
\begin{table}[htbp]
	\centering
	\caption{Performance comparison of SFM, DMFM and PD-DMFM under different amounts of high-fidelity data for the complex layout.}
	\resizebox{\textwidth}{!}{ 
		\begin{tabular}{ccllllllllll}
			\toprule
			\multicolumn{2}{c}{Amounts of high-fidelity data}  & 10    & 20    & 30    & 40    & 50    & 100   & 200   & 500   & 1000  & 2000 \\
			\midrule
			\multirow{3}{*}{SFM} & MAE   & 0.9723 & 0.4752 & 0.3548 & 0.3386 & 0.2727 & 0.195 & 0.1291 & 0.1153 & 0.0874 & 0.0725 \\
			& CMAE  & 0.592 & 0.3953 & 0.3175 & 0.3254 & 0.1991 & 0.1646 & 0.1152 & 0.1002 & 0.0786 & 0.0542 \\
			& MT-AE  & 5.1394 & 1.5032 & 0.4895 & 0.5598 & 0.4256 & 0.3974 & 0.1999 & 0.2001 & 0.1565 & 0.1133 \\
			\midrule
			\multirow{3}{*}{DMFM} & MAE   & 0.3491 & 0.2393 & \textbf{0.1644} & \textbf{0.1484} & \textbf{0.1296} & \textbf{0.1083} & \textbf{0.0963} & \textbf{0.0811} & \textbf{0.0778} & \textbf{0.066} \\
			& CMAE  & 0.2514 & 0.165 & \textbf{0.131} & \textbf{0.1185} & \textbf{0.1126} & \textbf{0.0967} & \textbf{0.0847} & \textbf{0.0705} & \textbf{0.0699} & 0.0568 \\
			& MT-AE  & 0.7241 & 0.3761 & 0.2525 & \textbf{0.2203} & \textbf{0.2115} & \textbf{0.1944} & \textbf{0.1369} & \textbf{0.1322} & \textbf{0.1169} & \textbf{0.0933} \\
			\midrule
			\multirow{3}{*}{PD-DMFM} & MAE   & \textbf{0.265} & \textbf{0.2191} & 0.2001 & 0.187 & 0.1854 & 0.1308 & 0.1038 & 0.0902 & 0.0828 & 0.068 \\
			& CMAE  & \textbf{0.1771} & \textbf{0.1612} & 0.1688 & 0.1429 & 0.1322 & 0.1128 & 0.0923 & 0.0845 & 0.0752 & \textbf{0.0564} \\
			& MT-AE  & \textbf{0.4363} & \textbf{0.3101} & \textbf{0.2393} & 0.2912 & 0.2572 & 0.2219 & 0.1693 & 0.1395 & 0.1242 & 0.1015 \\
			\bottomrule
	\end{tabular}}
	\label{tab:3}
\end{table}

\begin{figure}[htbp]
	\centering 
	\includegraphics[width=\textwidth]{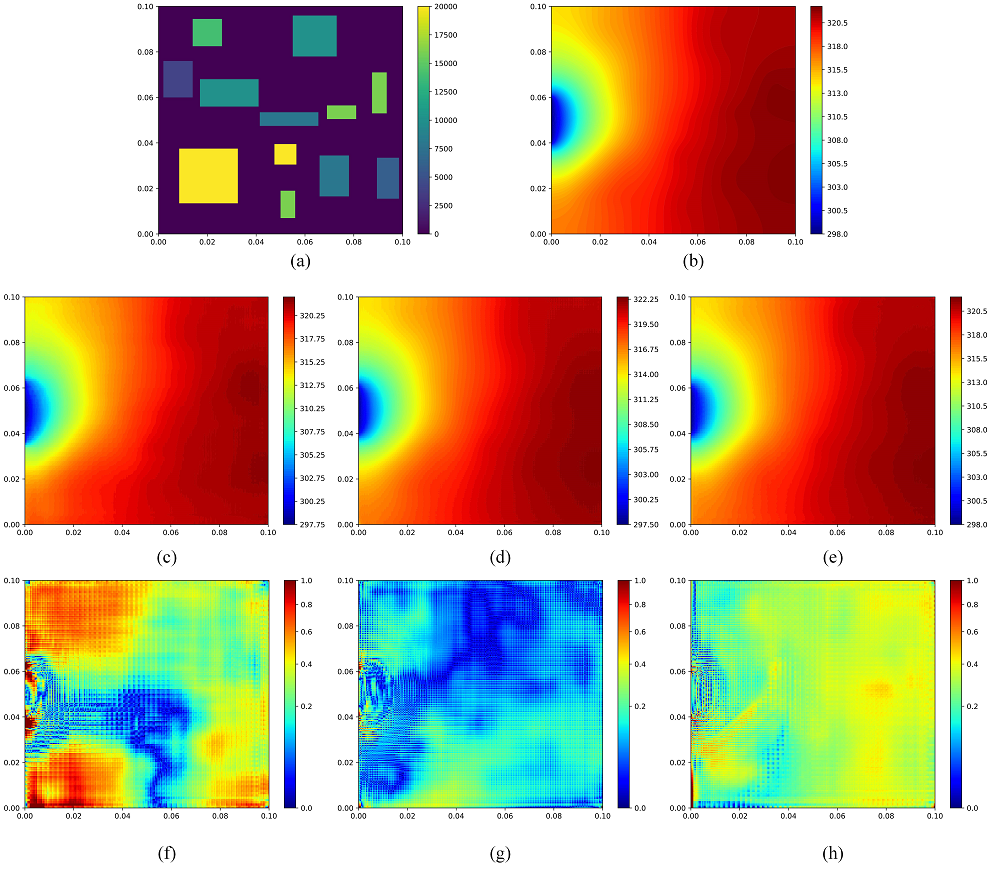}\\
	\caption{Visualization results for the complex layout. (a)complex layout, (b)ground-truth, (c)SFM prediction, (d)DMFM prediction, (e)PD-DMFM prediction, (f)prediction error of SFM, (g)prediction error of DMFM, (h)prediction error of PD-DMFM.}  
	\label{fig:10}
\end{figure}

The prediction of SFM, DMFM, and PD-DMFM trained by 50 high-fidelity samples is shown in Fig.\ref{fig:10}. It can be seen that the temperature predicted by SFM has a large deviation from the groud truth. The temperature field predicted by DMFM is basically consistent with the actual temperature, but there is a certain deviation around the heat emission hole. While the temperature prediction of PD-DMFM is the most accurate.
\begin{figure}[htbp] 
	\centering 
	\includegraphics[width=0.6\textwidth]{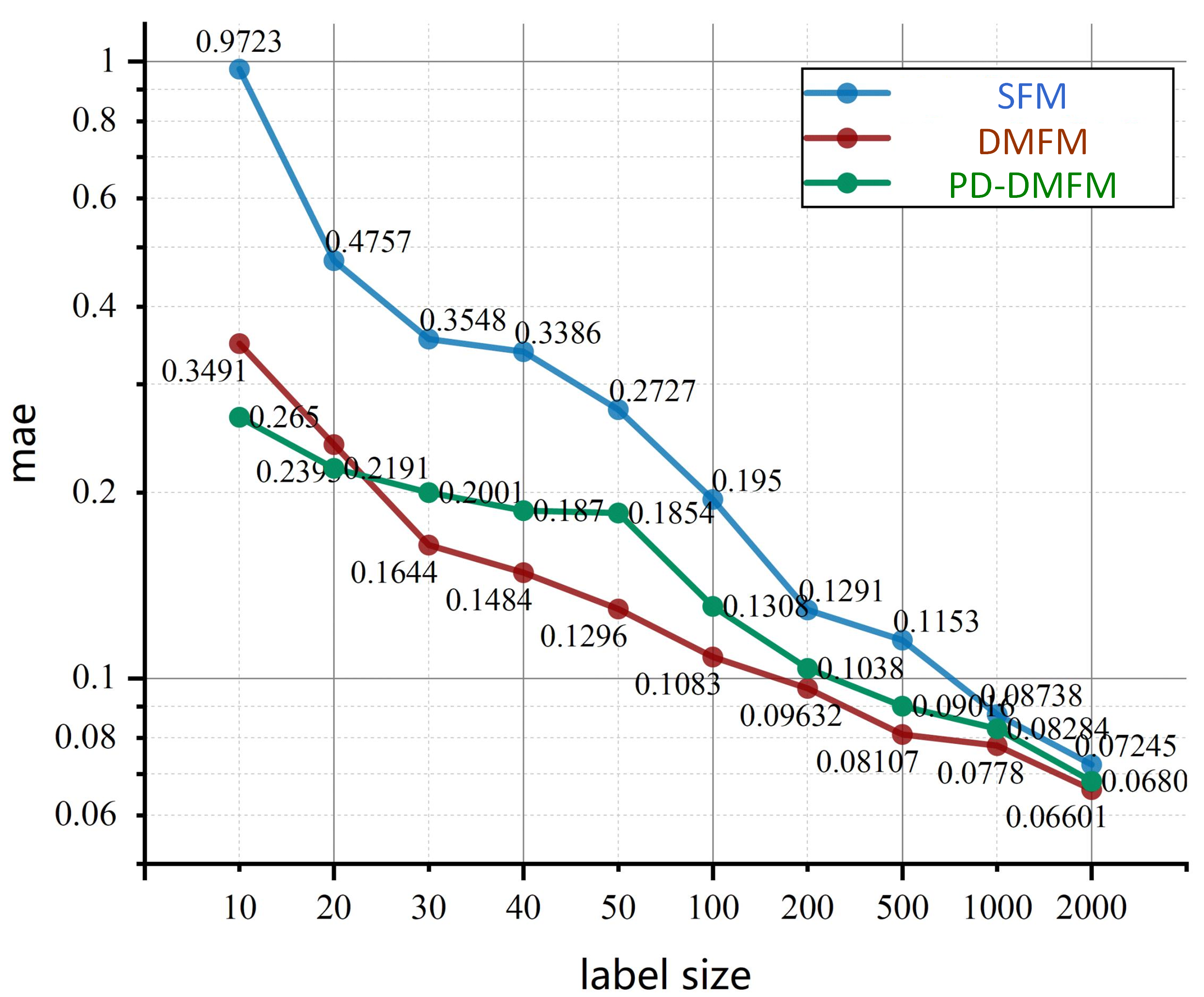}
	\caption{The changes of MAE with different amounts of high-fidelity data.}  
	\label{fig:9}
\end{figure}

The variation of MAE for temperature field prediction with different amounts of high-fidelity data is exhibited in Fig.\ref{fig:9}.
As observed in the simple layout, with the increase in high-fidelity data, the performance improvement of DMFM and PD-DMFM decreases. The phenomenon shows that DMFM and PD-DMFM proposed in this paper are more suitable for the situation with few high-fidelity data, improving the model's performance under limited high-fidelity data, which is consistent with our original intention to build a multi-fidelity model.

\section{Conclusions}
\label{sec:conclusions}
This paper proposes a novel deep multi-fidelity surrogate modeling method for temperature field prediction. The deep multi-fidelity model (DMFM) is based on the pre-train fine-tune paradigm. In the pre-train stage, the model first learns low-fidelity mapping through training with a large amount of low-fidelity data that contains inaccurate information. At this time, the model's prediction is not accurate but has learned useful features. The high-fidelity projection head replaces the low-fidelity projection head in the fine-tuning stage. Then,  the model is fine-tuned through limited high-fidelity data to complete the accurate high-fidelity mapping. Self-supervised learning based on the pre-train fine-tune paradigm is proposed further to reduce the reliance on the labeled low-fidelity data. 
A heat conduction equation with physical information is used to construct the loss function during the pre-train stage. Thus the physics-driven deep multi-fidelity model (PD-DMFM) is trained without many labeled low-fidelity data.
In addition, the temperature field prediction problem is constructed as an image-to-image regression task, and the proposed DMFM and PD-DMFM are tested with two various datasets. The result shows that the proposed method can greatly reduce the dependence of the model on high-fidelity data and save the construction cost of the model. The performance of DMFM  is the same as the single fidelity model trained by more high-fidelity data, and the PD-DMFM does not even need low-fidelity data to reach the same effect. Besides, the proposed method is general and can be extended to fluid mechanics and structural mechanics problems.

\section*{Acknowledgments}
This work was supported by the Postgraduate Scientific Research Innovation Project of Hunan Province (No.CX20200006) and the National Natural Science Foundation of China (Nos.11725211 and 52005505). 

\bibliography{mybibfile}

\begin{thebibliography}{10}
\expandafter\ifx\csname url\endcsname\relax
  \def\url#1{\texttt{#1}}\fi
\expandafter\ifx\csname urlprefix\endcsname\relax\def\urlprefix{URL }\fi
\expandafter\ifx\csname href\endcsname\relax
  \def\href#1#2{#2} \def\path#1{#1}\fi

\bibitem{Yao2011}
W.~Yao, X.~Chen, W.~Luo, M.~van Tooren, J.~Guo, Review of uncertainty-based
  multidisciplinary design optimization methods for aerospace vehicles,
  Progress in Aerospace Sciences 47~(6) (2011) 450--479.
\newblock \href {http://dx.doi.org/10.1016/j.paerosci.2011.05.001}
  {\path{doi:10.1016/j.paerosci.2011.05.001}}.

\bibitem{Zheng2021}
W.~Yao, X.~Zheng, J.~Zhang, N.~Wang, G.~Tang, Deep adaptive arbitrary
  polynomial chaos expansion: A mini-data-driven semi-supervised method for
  uncertainty quantification (2021).
\newblock \href {http://dx.doi.org/10.48550/ARXIV.2107.10428v2}
  {\path{doi:10.48550/ARXIV.2107.10428v2}}.

\bibitem{Zheng2019}
X.~Zheng, W.~Yao, Y.~Xu, X.~Chen, Improved compression inference algorithm for
  reliability analysis of complex multistate satellite system based on
  multilevel bayesian network, Reliability Engineering \& System Safety 189
  (2019) 123--142.
\newblock \href {http://dx.doi.org/10.1016/j.ress.2019.04.011}
  {\path{doi:10.1016/j.ress.2019.04.011}}.

\bibitem{Zheng2020}
X.~Zheng, W.~Yao, Y.~Xu, X.~Chen, Algorithms for bayesian network modeling and
  reliability inference of complex multistate systems: Part i – independent
  systems, Reliability Engineering \& System Safety 202.
\newblock \href {http://dx.doi.org/10.1016/j.ress.2020.107011}
  {\path{doi:10.1016/j.ress.2020.107011}}.

\bibitem{he2021thermal}
Z.~He, Y.~Yan, Z.~Zhang, Thermal management and temperature uniformity
  enhancement of electronic devices by micro heat sinks: A review, Energy 216
  (2021) 119223.

\bibitem{ahmed2018optimization}
H.~E. Ahmed, B.~Salman, A.~S. Kherbeet, M.~Ahmed, Optimization of thermal
  design of heat sinks: A review, International Journal of Heat and Mass
  Transfer 118 (2018) 129--153.

\bibitem{goel2009comparing}
T.~Goel, R.~T. Hafkta, W.~Shyy, Comparing error estimation measures for
  polynomial and kriging approximation of noise-free functions, Structural and
  Multidisciplinary Optimization 38~(5) (2009) 429.

\bibitem{clark2016engineering}
D.~L. Clark~Jr, H.-R. Bae, K.~Gobal, R.~Penmetsa, Engineering design
  exploration using locally optimized covariance kriging, AIAA Journal 54~(10)
  (2016) 3160--3175.

\bibitem{yao2011concurrent}
W.~Yao, X.~Chen, Y.~Zhao, M.~van Tooren, Concurrent subspace width optimization
  method for rbf neural network modeling, IEEE transactions on neural networks
  and learning systems 23~(2) (2011) 247--259.

\bibitem{clarke2005analysis}
S.~M. Clarke, J.~H. Griebsch, T.~W. Simpson, Analysis of support vector
  regression for approximation of complex engineering analyses.

\bibitem{lecun2015deep}
Y.~LeCun, Y.~Bengio, G.~Hinton, Deep learning, nature 521~(7553) (2015)
  436--444.

\bibitem{chen2021deep}
X.~Chen, X.~Zhao, Z.~Gong, J.~Zhang, W.~Zhou, X.~Chen, W.~Yao, A deep neural
  network surrogate modeling benchmark for temperature field prediction of heat
  source layout, arXiv preprint arXiv:2103.11177.

\bibitem{zhao2021physics}
X.~Zhao, Z.~Gong, Y.~Zhang, W.~Yao, X.~Chen, Physics-informed convolutional
  neural networks for temperature field prediction of heat source layout
  without labeled data, arXiv preprint arXiv:2109.12482.

\bibitem{chen2020heat}
X.~Chen, X.~Chen, W.~Zhou, J.~Zhang, W.~Yao, The heat source layout
  optimization using deep learning surrogate modeling, Structural and
  Multidisciplinary Optimization 62~(6) (2020) 3127--3148.

\bibitem{zhao2021surrogate}
X.~Zhao, Z.~Gong, J.~Zhang, W.~Yao, X.~Chen, A surrogate model with data
  augmentation and deep transfer learning for temperature field prediction of
  heat source layout, Structural and Multidisciplinary Optimization 64~(4)
  (2021) 2287--2306.

\bibitem{gong2019cnn}
Z.~Gong, P.~Zhong, Y.~Yu, W.~Hu, S.~Li, A cnn with multiscale convolution and
  diversified metric for hyperspectral image classification, IEEE Transactions
  on Geoscience and Remote Sensing 57~(6) (2019) 3599--3618.

\bibitem{gong2020statistical}
Z.~Gong, P.~Zhong, W.~Hu, Statistical loss and analysis for deep learning in
  hyperspectral image classification, IEEE Transactions on Neural Networks and
  Learning Systems 32~(1) (2020) 322--333.

\bibitem{peherstorfer2018survey}
B.~Peherstorfer, K.~Willcox, M.~Gunzburger, Survey of multifidelity methods in
  uncertainty propagation, inference, and optimization, Siam Review 60~(3)
  (2018) 550--591.

\bibitem{fernandez2016review}
M.~G. Fern{\'a}ndez-Godino, C.~Park, N.-H. Kim, R.~T. Haftka, Review of
  multi-fidelity models, arXiv preprint arXiv:1609.07196.

\bibitem{le2014recursive}
L.~Le~Gratiet, J.~Garnier, Recursive co-kriging model for design of computer
  experiments with multiple levels of fidelity, International Journal for
  Uncertainty Quantification 4~(5).

\bibitem{perdikaris2015multi}
P.~Perdikaris, D.~Venturi, J.~O. Royset, G.~E. Karniadakis, Multi-fidelity
  modelling via recursive co-kriging and gaussian--markov random fields,
  Proceedings of the Royal Society A: Mathematical, Physical and Engineering
  Sciences 471~(2179) (2015) 20150018.

\bibitem{bierig2016approximation}
C.~Bierig, A.~Chernov, Approximation of probability density functions by the
  multilevel monte carlo maximum entropy method, Journal of Computational
  Physics 314 (2016) 661--681.

\bibitem{giles2008multilevel}
M.~B. Giles, Multilevel monte carlo path simulation, Operations research 56~(3)
  (2008) 607--617.

\bibitem{liu2019multi}
D.~Liu, Y.~Wang, Multi-fidelity physics-constrained neural network and its
  application in materials modeling, Journal of Mechanical Design 141~(12).

\bibitem{zhang2021multi}
X.~Zhang, F.~Xie, T.~Ji, Z.~Zhu, Y.~Zheng, Multi-fidelity deep neural network
  surrogate model for aerodynamic shape optimization, Computer Methods in
  Applied Mechanics and Engineering 373 (2021) 113485.

\bibitem{chen2022multi}
J.~Chen, Y.~Gao, Y.~Liu, Multi-fidelity data aggregation using convolutional
  neural networks, Computer Methods in Applied Mechanics and Engineering 391
  (2022) 114490.

\bibitem{song2021transfer}
D.~H. Song, D.~M. Tartakovsky, Transfer learning on multi-fidelity data, arXiv
  preprint arXiv:2105.00856.

\bibitem{meng2020composite}
X.~Meng, G.~E. Karniadakis, A composite neural network that learns from
  multi-fidelity data: Application to function approximation and inverse pde
  problems, Journal of Computational Physics 401 (2020) 109020.

\bibitem{chakraborty2021transfer}
S.~Chakraborty, Transfer learning based multi-fidelity physics informed deep
  neural network, Journal of Computational Physics 426 (2021) 109942.

\bibitem{chen2016optimization}
K.~Chen, S.~Wang, M.~Song, Optimization of heat source distribution for
  two-dimensional heat conduction using bionic method, International Journal of
  Heat and Mass Transfer 93 (2016) 108--117.

\bibitem{chen2017heat}
K.~Chen, J.~Xing, S.~Wang, M.~Song, Heat source layout optimization in
  two-dimensional heat conduction using simulated annealing method,
  International Journal of heat and Mass transfer 108 (2017) 210--219.

\bibitem{aslan2018heat}
Y.~Aslan, J.~Puskely, A.~Yarovoy, Heat source layout optimization for
  two-dimensional heat conduction using iterative reweighted l1-norm convex
  minimization, International Journal of Heat and Mass Transfer 122 (2018)
  432--441.

\bibitem{vapnik1992principles}
V.~Vapnik, Principles of risk minimization for learning theory, in: Advances in
  neural information processing systems, 1992, pp. 831--838.

\bibitem{raissi2019physics}
M.~Raissi, P.~Perdikaris, G.~E. Karniadakis, Physics-informed neural networks:
  A deep learning framework for solving forward and inverse problems involving
  nonlinear partial differential equations, Journal of Computational Physics
  378 (2019) 686--707.

\bibitem{he2015delving}
K.~He, X.~Zhang, S.~Ren, J.~Sun, Delving deep into rectifiers: Surpassing
  human-level performance on imagenet classification, in: Proceedings of the
  IEEE international conference on computer vision, 2015, pp. 1026--1034.

\bibitem{Ranger}
L.~Wright, Ranger - a synergistic optimizer.,
  \url{https://github.com/lessw2020/Ranger-Deep-Learning-Optimizer} (2019).

\bibitem{liu2019variance}
L.~Liu, H.~Jiang, P.~He, W.~Chen, X.~Liu, J.~Gao, J.~Han, On the variance of
  the adaptive learning rate and beyond, arXiv preprint arXiv:1908.03265.

\bibitem{zhang2019lookahead}
M.~R. Zhang, J.~Lucas, G.~Hinton, J.~Ba, Lookahead optimizer: k steps forward,
  1 step back, arXiv preprint arXiv:1907.08610.

\bibitem{loshchilov2016sgdr}
I.~Loshchilov, F.~Hutter, Sgdr: Stochastic gradient descent with warm restarts,
  arXiv preprint arXiv:1608.03983.

\bibitem{paszke2019pytorch}
A.~Paszke, S.~Gross, F.~Massa, A.~Lerer, J.~Bradbury, G.~Chanan, T.~Killeen,
  Z.~Lin, N.~Gimelshein, L.~Antiga, et~al., Pytorch: An imperative style,
  high-performance deep learning library, Advances in neural information
  processing systems 32 (2019) 8026--8037.

\end{thebibliography}

\end{document}